# SP-Guide: Enhancing Aerial Vision-Language Navigation with Spatial Priors

Haoxiang Tang
*School of Astronautics, Harbin Institute of Technology*
Harbin, 150001, China
25S004022@stu.hit.edu.cn

Dong Zhou
*School of Astronautics, Harbin Institute of Technology*
Harbin, 150001, China
dongzhou@hit.edu.cn

Shen Tan
*School of Astronautics, Harbin Institute of Technology*
Harbin, 150001, China
shentan@stu.hit.edu.cn

Minxuan Jiang
*School of Astronautics, Harbin Institute of Technology*
Harbin, 150001, China
Alphal Uohn@gmail.com

Xitong Li
*The 28th Research Institute, China Electronics Technology Group Corporation*
Nanjing, 210001, China
545005922@qq.com

***Abstract*—Aerial vision-language navigation (AVLN) requires a UAV to reach a language-described target from partial egocentric observations while reasoning over large outdoor scenes. Existing approaches typically fuse view-centered RGB observations and globally anchored history maps as generic visual embeddings, leaving cross-modal spatial alignment to be learned from sparse navigation supervision. We present SP-Guide, a transformer-based goal-prediction framework that addresses this weakness of alignment through a Spatial Prior Cross-Modal Encoder, an Instruction Enhanced Decoder, and a Goal Convergence Filter for terminal target refinement. Experiments implemented on the CityNav benchmark show that SP-Guide improves success rate over the representative baselines, with the largest gains on unseen environments. These results indicate that physically grounded spatial priors provide an effective inductive bias for robust AVLN. The code is available at https://github.com/Dongzhou-1996/SP-Guide.**

## I. Introduction

In the Aerial vision-language navigation (AVLN) task, an unmanned aerial vehicle navigates based on natural-language guidance toward a target while perceiving the world through partial egocentric observations. This capability is relevant to urban inspection, emergency response, and low-altitude autonomous exploration, where an aerial agent must repeatedly connect local evidence to large-scale scene structure [1], [2].

Compared with indoor Vision-and-Language Navigation (VLN), the aerial setting introduces substantially greater difficulties. First, viewpoints shift rapidly during flight, leading to limited visual overlap between consecutive frames and requiring the agent to robustly handle large motion discontinuities. Second, target objects are often distant and thus occupy only a small portion of the image, which demands fine-grained feature extraction and multi-scale perception to reliably detect and localise them. Third, long-range spatial consistency must be maintained throughout the entire trajectory, as the agent needs to continuously integrate local observations into a coherent global representation for accurate path planning and goal reaching.

A central difficulty in AVLN is multimodal spatial mismatch. The agent reasons jointly over egocentric RGB/depth observations and a globally anchored semantic map. These modalities are complementary, but they do not share the same coordinate semantics: visual observation embeddings are viewcentered, whereas global map embeddings are world-centered. If they are fused as generic modality embeddings, the network must infer cross-modal correspondence, instruction relevance, and goal geometry simultaneously from sparse supervision. This ambiguity becomes especially severe when several regions are semantically plausible but only one is consistent with both the instruction and the global layout.

Terminal decision making introduces a second source of error. In goal-prediction agents [2], the predicted progress score is often used as an approximate stop signal. When the progress estimate is noisy, the agent may stop too early, overshoot the target, or oscillate around the goal region. Consequently, better one-step prediction does not automatically translate into stronger closed-loop navigation.

To address these issues mentioned above, we present the **SP-Guide** method, a transformer-based AVLN framework that introduces spatially constrained cross-modal reasoning and more reliable terminal commitment. The method preserves standard preprocessing and pretrained modality encoders, but strengthens how local observations are aligned with the global map and how the final target is selected near the end of rollout.

This design keeps the comparison focused on representation and decision quality rather than on broad architectural replacement.

Our contributions are summarized as follows:

- We propose **SP-Guide,** an aerial vision-language navigation framework that enhances transformer-based goal prediction with **spatial priors.**
- We introduce the **Goal Convergence Filter (GCF),** a lightweight terminal reasoning module that aggregates recent goal and progress predictions to produce a more reliable final target and stopping decision.
- Experimental results show that SP-Guide consistently outperforms conventional baselines on a large-scale aerial vision-language navigation benchmark, demonstrating the value of spatial-prior-guided reasoning for robust navigation.

## II. Related Work

### A. Aerial Vision-Language Navigation

Vision-language navigation first gained traction in graph-structured indoor environments, where the agent typically moves over a discrete navigation graph and follows long natural-language route instructions [3]-[5], This formulation enabled rapid progress in instruction grounding, progress estimation, and imitation learning, but it also hides many low-level control difficulties behind oracle graph transitions. Continuous indoor VLN therefore emerged as an important next step. VLN-CE removes the navigation graph and requires agents to act in continuous 3D environments from RGB-D observations, establishing *Seq2Seq* and *CMA* as standard embodied baselines [6]. Later work explored language-conditioned waypoint prediction [7] and explicit cross-modal map learning [8], showing that continuous VLN performance depends strongly on how local perception, spatial memory, and language are coupled. These indoor studies are important references for AVLN because they expose the same representation-to-action bottleneck, even though aerial navigation involves larger outdoor scenes and more severe viewpoint changes.

Recent aerial benchmarks extend this line of work to geographically large outdoor scenes [1], [2], CityNav is especially relevant because it provides a large-scale real-world urban AVLN dataset with natural-language descriptions, human demonstration trajectories, and a geographic semantic map (GSM) that encodes stable top-down spatial cues and instruction-mentioned landmarks. This landmark-aware map representation improves long-range exploration efficiency by giving the agent geographically meaningful anchors rather than relying only on short-range egocentric evidence.

### B. Spatial Priors in Vision Transformers

The Transformer was originally introduced for sequence transduction with multi-head self-attention and encoder-decoder processing [9], Its success motivated direct adaptation to vision, where ViT treats image patches as tokens and processes them with a standard Transformer encoder [10]. Subsequent variants improved the visual inductive bias of this design: DeiT made ViT training more data efficient through attention-based distillation [11], while Swin Transformer introduced hierarchical shifted-window attention to recover locality and multi-scale structure [12]. These developments show that although attention provides flexible global interaction, effective visual transformers often still require spatial structure to be injected through tokenization, locality, hierarchy, or positional bias.

A more focused line of work therefore reintroduces geometry directly into token interaction. CPVT [13] replaces fixed positional embeddings with conditional positional encodings, while SP-ViT [14] and enhanced-spatial-prior attention modules explicitly bias token interaction with two-dimensional geometry [15]. Together, these studies show that transformer attention benefits from spatial priors rather than relying entirely on feature similarity.

Most spatial-prior transformers are designed for a single visual lattice. Their priors are typically defined over patch neighborhoods or relative offsets within one modality. That assumption is appropriate for image classification and dense prediction, but it is incomplete for AVLN with semantic maps. Our setting requires cross-modal grounding across two coordinate systems: a view-centered observation lattice and a globally anchored map lattice. SP-Guide extends spatial-prior reasoning from *within-modality locality* to *cross-modality spatial consistency* by constraining attention with distances measured in a shared physical frame.

## III. Method

### A. Task Formulation

We consider aerial vision-language navigation (AVLN) in a large-scale urban environment. At the beginning of each episode, a UAV agent receives a natural-language instruction $L$ that describes a target location or destination. The agent starts from an initial pose $(p_0, \psi_0)$. where $p_0 \in \mathbb{R}^2$ denotes the planar position and $\psi_0$ denotes the heading angle. During navigation, the agent does not observe the full environment; at each step $t$, it receives a partial observation $o_t$ and selects a discrete action $a_t \in \mathcal{A}$ conditioned on the instruction and observation history.

Let $g^* \in \mathbb{R}^2$ denote the ground-truth target position in the global coordinate frame. Executing the policy for at most $T$ steps produces a trajectory of poses, observations, and actions. The objective is to generate an action sequence that follows the instruction and drives the UAV to $g^*$. An episode is considered successful if the final UAV position lies within a predefined tolerance radius $\epsilon$ of the target.

### B. SP-Guide Overview

As shown in Fig. 1, SP-Guide consists of three main parts: a *Spatial Prior Cross-Modal Encoder* (SPCE), an *Instruction Enhanced Decoder* (IED), and a *Goal Convergence Filter* (GCF). SPCE builds a shared visual memory from multimodal observations while injecting spatial constraints between local observation tokens and global map tokens. IED uses the instruction to query this visual memory and extract navigation-relevant context. From the decoded representation, the model predicts a goal estimate $\hat{\mathbf{g}}$ and a progress estimate $\hat{p}$. These predictions are converted into actions by a goal-directed selector, while GCF refines the final target and stopping decision near the end of rollout.

SP-Guide adopts the geographic semantic map (GSM) representation introduced by CityNav [2]. This map provides a compact global reference that complements egocentric RGB and depth observations, allowing the agent to reason jointly over local appearance and large-scale spatial layout.

For feature construction, SP-Guide uses pretrained modality encoders to extract token embeddings from language, RGB, and depth. We denote the instruction tokens by $E_L$, and the RGB, depth, and map tokens by $E_r$, $E_d$, and $E_m$. These visual tokens are organized into a unified visual sequence and passed into SPCE, while $E_L$ is preserved as the language-side representation queried by IED.

### C. Spatial Prior Cross-Modal Encoder

All visual tokens are first processed by the *Spatial Prior Cross-Modal Encoder* (SPCE). Its key role is to regularize cross-modal interaction with an explicit physical prior instead of forcing the model to infer RGB-map correspondence only from sparse endpoint supervision. Although RGB, depth, and map tokens all pass through the encoder, the core design focuses on the observation-map branch, where the spatial mismatch is most severe.

Figure 2 shows the computation graph of the proposed spatial-prior encoder. The design keeps the transformer block simple: content attention is first computed in the usual way, softly modulated by a distance-derived cross-modal prior, and then passed through residual and feed-forward updates. This decomposition cleanly separates standard content attention from the proposed spatial modulation.

Let $E \in \mathbb{R}^{l\times d}$ denote the concatenated visual-token sequence, where $l$ is the number of visual tokens and $d$ is the hidden dimension. The $i$-th token in $E$ is associated with a projected patch center $\mathbf{c}_i \in [0,1]^2$. For map tokens, $\mathbf{c}_i$ is given by the corresponding map-grid patch center. For observation tokens, $\mathbf{c}_i$ is obtained by projecting the image patch center into the current observation footprint centered at the UAV position $\mathbf{p}$. This provides a shared normalized physical frame for local observations and the global map.

From the visual sequence $E$, we compute standard content attention:

$$Q = EW_Q, \qquad K = EW_K, \qquad V = EW_V, \tag{1}$$

$$A_{ij} = \mathrm{softmax}\left(\frac{Q_i K_j^\top}{\sqrt{d}}\right), \tag{2}$$

where $A_{ij}$ is the content-based attention between the $i$-th query token and the $j$-th key token. We then build the patch-distance matrix

$$D_{ij} = \left\|\mathbf{c}_i - \mathbf{c}_j\right\|_2. \tag{3}$$

The distance matrix is converted into a spatial prior through

$$S_{ij} = \exp(-D_{ij}/\tau), \tag{4}$$

where $\tau$ is a learned positive scale. To avoid over-sharpening the prior, we apply a *soft-reweight* step:

$$\tilde{S}_{ij} = \frac{1}{2}\left(S_{ij} + \frac{1}{l}\sum_{k=1}^{l} S_{ik}\right). \tag{5}$$

The attention matrix is then spatially modulated as

$$A'_{ij} = A_{ij}\Big(1 + \alpha(\tilde{S}_{ij} - 1)\Big), \tag{6}$$

where $\alpha$ is a learnable parameter to control the strength of the spatial modulation. The updated token sequence is

$$E_i^{\mathrm{sp}} = \sum_{j=1}^{l} A'_{ij} V_j. \tag{7}$$

Crucially, SPCE does not define the prior only in image space. Instead, it modulates attention with distances measured in a shared physical frame, encouraging the model to align local observations and global map regions that are physically compatible rather than merely visually similar. Depth tokens remain in the same visual sequence and benefit from the globally informed context, while the explicit prior is concentrated on the cross-modal interaction that carries the main geometric burden.

### D. Instruction Enhanced Decoder

After SPCE, the spatially refined visual sequence $E^{\mathrm{sp}}$ is encoded into a visual memory $H = \mathrm{Enc}(E^{\mathrm{sp}})$. The *Instruction Enhanced Decoder* then reads task-relevant content from this memory. The decoder query is formed by prepending a learnable navigation token $q_0$ to the instruction tokens, i.e., $Q_L = [q_0; E_L]$, and the decoded representation is obtained as $Z = \mathrm{Dec}(Q_L, H)$. We use the first decoded token $z = Z_0$ as the navigation summary, from which a goal head $f_g(\cdot)$ and a progress head $f_p(\cdot)$ produce the goal estimate $\hat{\mathbf{g}}_t$ and the progress estimate $\hat{p}_t$.

This design avoids treating language as passive context inside an encoder-only pool. Instead, instruction tokens actively query the visual memory, while the learnable token provides a compact aggregation anchor for final prediction. Compared with decoder variants that rely on a free learnable query alone, IED keeps the query instruction-aware and therefore better aligned with target-relevant evidence.

### E. Goal Convergence Filter

Even with a stronger predictor, final navigation quality can still be degraded by a brittle stop rule. We therefore add the *Goal Convergence Filter* (GCF) as a lightweight terminal reasoning module. Let the last $K$ predicted goals and progress values be

$$\{(\hat{\mathbf{g}}_i, \hat{p}_i)\}_{i=t-K+1}^{t}. \tag{8}$$

We first convert progress into normalized weights

$$w_i = \frac{\exp(\beta\hat{p}_i)}{\sum_{j=t-K+1}^{t}\exp(\beta\hat{p}_j)}, \tag{9}$$

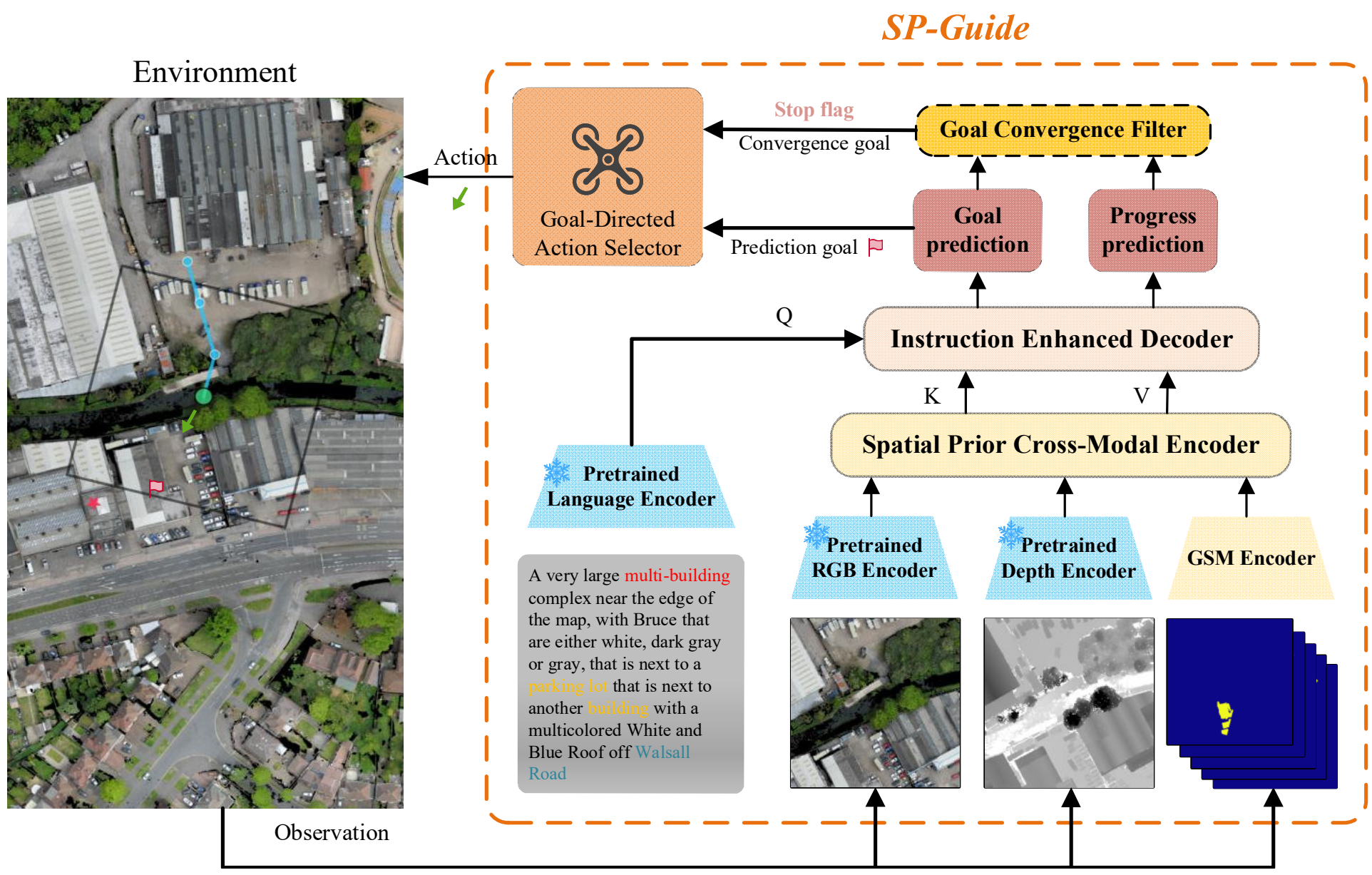

Fig. 1. Overview of **SP-Guide**. SPCE aligns local observations with the global map, IED extracts instruction-conditioned navigation context, and GCF refines the terminal goal and stopping decision during rollout.

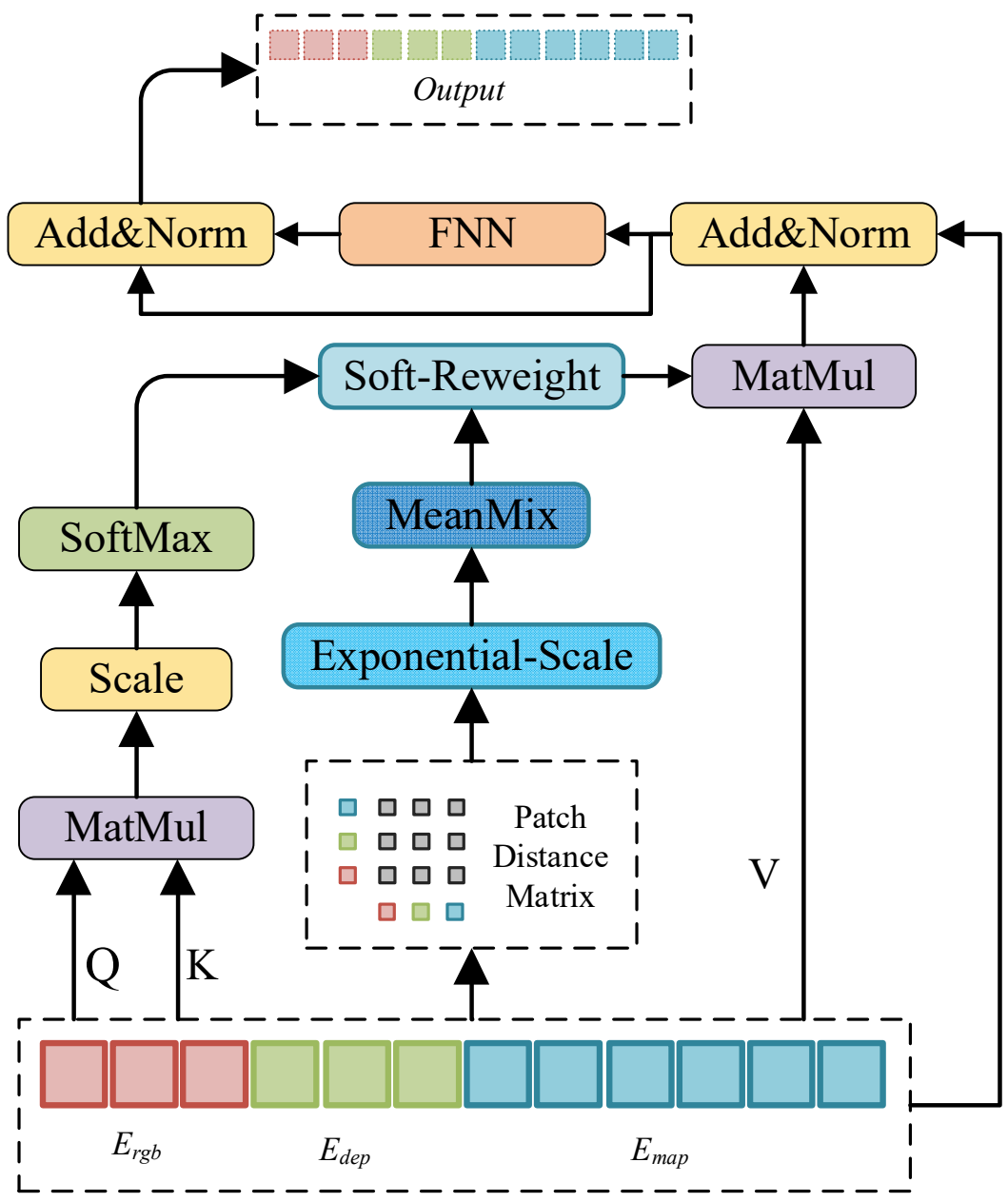

Fig. 2. Structure of the **Spatial Prior Cross-Modal Encoder**. Observation-map interaction is modulated by a patch-distance prior in a normalized physical frame.

where $\beta$ controls how strongly high-progress predictions are emphasized. GCF estimates the terminal goal by solving the weighted least-squares problem

$$\bar{\mathbf{g}}_t = \arg\min_{\mathbf{g}} \sum_{i=t-K+1}^{t} w_i \left\|\mathbf{g} - \hat{\mathbf{g}}_i\right\|_2^2 . \tag{10}$$

This objective has the closed-form solution

$$\bar{\mathbf{g}}_t = \sum_{i=t-K+1}^{t} w_i \hat{\mathbf{g}}_i . \tag{11}$$

We then compute the weighted convergence statistic

$$v^\star = \sum_{i=t-K+1}^{t} w_i \left\|\hat{\mathbf{g}}_i - \bar{\mathbf{g}}_t\right\|_2^2 . \tag{12}$$

We also compute the weighted terminal progress

$$\bar{p}_t = \sum_{i=t-K+1}^{t} w_i \hat{p}_i . \tag{13}$$

If $v^\star$ falls below a threshold $\tau_v$ and $\bar{p}_t$ exceeds a threshold $\tau_p$, the terminal window is treated as converged and $\bar{\mathbf{g}}_t$ is locked as the final target; otherwise the rollout continues. The locked target is passed to the goal-directed action selector, which converts it into the nearest feasible discrete motion command.

## IV. Experiments

### A. Setup

We evaluate our method on the CityNav benchmark [2]. Following the benchmark protocol, we report results on val-seen, val-unseen, and test-unseen. The first two splits are used for model development and parameter selection, while test-unseen is used only for final reporting with all settings fixed. The main navigation metrics are navigation error (NE), success rate (SR), oracle success rate (OSR), and success weighted by path length (SPL). For SP-Guide, GCF is applied online during rollout with a terminal window of $K = 3$, inverse temperature $\beta = 4.0$, variance threshold $\tau_v = 1400$, minimum lock step of 7, and arrival radius of 5 meters. If no terminal window satisfies the convergence criterion, the agent continues normal

TABLE I
QUANTITATIVE RESULTS ON THE CITYNAV DATASET.

| Models | Validation Seen | | | | Validation Unseen | | | | Test Unseen | | | |
|---|---|---|---|---|---|---|---|---|---|---|---|---|
| | NE↓ | SR↑ | OSR↑ | SPL↑ | NE↓ | SR↑ | OSR↑ | SPL↑ | NE↓ | SR↑ | OSR↑ | SPL↑ |
| Random [2] | 222.3 | 0.00 | 1.15 | 0.00 | 223.0 | 0.00 | 0.90 | 0.00 | 208.8 | 0.00 | 1.44 | 0.00 |
| Human [2] | 9.1 | 89.31 | 96.40 | 60.17 | 9.4 | 88.39 | 95.54 | 62.66 | 9.8 | 87.86 | 95.29 | 57.04 |
| Seq2Seq [2] | 257.1 | 1.81 | 7.89 | 1.58 | 317.4 | 0.79 | 8.82 | 0.61 | 245.3 | 1.50 | 8.34 | 1.30 |
| CMA [2] | 240.8 | 0.95 | 9.42 | 0.92 | 268.8 | 0.65 | 7.86 | 0.63 | 252.6 | 0.82 | 9.70 | 0.79 |
| AerialVLN [1] | 65.6 | 9.77 | 23.77 | 8.64 | 81.8 | 6.79 | 17.91 | 5.73 | **64.1** | 8.09 | 19.13 | 5.91 |
| MGP [2] | 56.7 | 14.93 | 22.54 | 13.90 | 76.9 | 6.55 | 14.19 | 6.09 | 86.0 | 10.05 | 17.74 | 9.17 |
| **SP-Guide** | **54.5** | **17.37** | **33.47** | **16.34** | **64.6** | **12.49** | **27.28** | **11.66** | 84.8 | **14.39** | **29.22** | **13.45** |

goal-directed rollout until the horizon. The success radius is set to 20 meters.

For training, we use the human-collected trajectories. The supervision target contains the normalized ground-truth goal location $\tilde{\mathbf{g}}^\star$ and a scalar progress label $y_t$, defined as the clipped fraction of distance-to-goal reduction from the start position $\mathbf{p}_0$ to the current position $\mathbf{p}_t$. All proposed variants are trained for 10 epochs with AdamW, using a learning rate of $1.5\times10^{-3}$ and a batch size of 12. The training objective combines goal regression and progress regression with equal mean-squared-error weights. The training-loss curves are omitted here to keep the paper within the page limit; the main takeaway is that the spatial-prior setting improves goal localization more than progress supervision.

### *B. Main Results*

The main comparison follows the CityNav reporting style and reports all three official splits. The baseline rows are taken from the CityNav reference results, while SP-Guide is evaluated with the same online GCF backend used throughout our ablation study. SPL is reported under the corresponding rollout traces.

Table I shows that SP-Guide improves final navigation over the reported CityNav baselines under the same metric definitions. Compared with MGP, SP-Guide improves SR by $2.4\%$ on *val-seen*, $6.9\%$ on *val-unseen*, and $4.3\%$ on *test-unseen*. It also improves OSR and SPL on all three splits, indicating that the spatial-prior-guided representation and terminal GCF improve closed-loop navigation rather than merely one-step goal prediction. However, the NE of AerialVLN is slightly better than SP-Guide on *test-unseen*, indicating that the spatial prior may not be sufficient for coarse-grained navigation in unseen environments. Figure 3 provides a qualitative example in which SP-Guide progressively approaches the target and terminates near the predicted goal region.

### *C. Ablations and Discussion*

The ablation in Table II evaluates four variants of our framework: the base decoder, the decoder augmented with spatial priors, the decoder with GCF, and the full SP-Guide model. The first two use a fixed-threshold progress-based stopping rule, while the latter two employ the online terminal convergence filter described above.

The ablation confirms that terminal reasoning is critical for converting one-step goal predictions into final navigation success. GCF consistently improves the base IED model across all splits, for example raising SR from 12.89 to 16.17 on *val-seen* and from 10.79 to 15.31 on *test-unseen*. Spatial priors provide additional gains on the validation splits when paired with GCF, where SP-Guide achieves the best SR and SPL on both *val-seen* and *val-unseen*. On *test-unseen*, the GCF variant is slightly stronger, suggesting that the spatial prior improves validation generalization but still leaves room for more robust cross-scene calibration.

The fact that SP-Guide is not always the strongest on *test-unseen* is also informative: a shared normalized physical frame is useful, but not yet sufficient to capture all cross-scene variation in outdoor navigation, especially when the map distribution or landmark arrangement shifts.

### *D. Limitations and Future Work*

SP-Guide remains a lightweight approximation of cross-modal geometry rather than a full geometric reconstruction system. SPCE aligns observation patches and map patches through a shared normalized frame induced by the current UAV footprint, which is practical but still lossy. Likewise, GCF is a lightweight terminal reasoning module rather than a fully learned uncertainty-aware stopping policy. Future work could add richer world-coordinate grounding, uncertainty-aware goal distributions, and stronger agent-level training beyond supervised goal regression.

## V. CONCLUSION

We propose SP-Guide, an AVLN framework that integrates three key components: instruction-enhanced decoding, physical-space spatial priors, and terminal convergence filtering. Experimental results on the CityNav benchmark show that SP-Guide achieves the overall best performance, surpassing conventional methods such as Seq2Seq, CMA and MGP. To overcome the remaining limitation of poor generalization in unseen scenarios, our future work will focus on efficient data augmentation techniques and advanced training paradigms to further improve the robustness and practicality of the system.

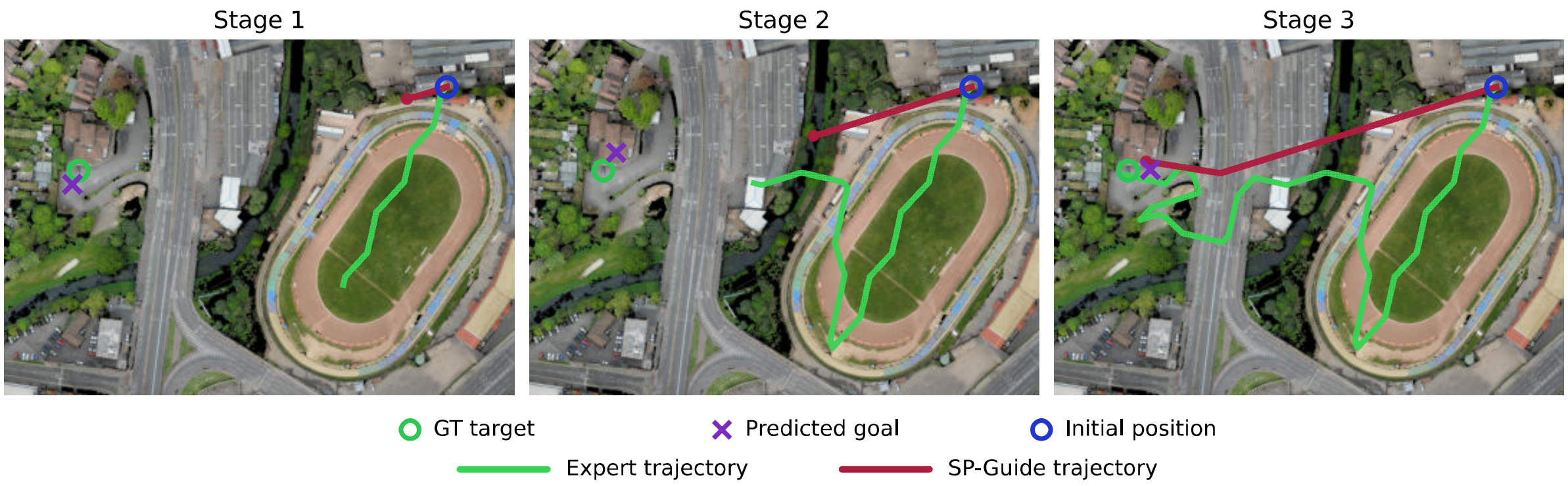

Fig. 3. Qualitative rollout of SP-Guide on a successful *val-unseen* episode. The expert path is shown in green, the SP-Guide trajectory is divided into three stages, and each panel marks the corresponding predicted goal.

TABLE II
ABLATED RESULTS OF THE MAIN COMPONENTS ON THE CITYNAV DATASET.

| # | SP | GCF | Validation Seen | | | | Validation Unseen | | | | Test Unseen | | | |
|---|---|---|---|---|---|---|---|---|---|---|---|---|---|---|
| | | | NE↓ | SR↑ | OSR↑ | SPL↑ | NE↓ | SR↑ | OSR↑ | SPL↑ | NE↓ | SR↑ | OSR↑ | SPL↑ |
| 1 | - | - | 56.0 | 12.89 | 25.02 | 12.13 | 78.1 | 7.22 | 18.40 | 6.72 | 83.9 | 10.79 | 22.75 | 10.08 |
| 2 | ✓ | - | 60.8 | 12.99 | 23.34 | 11.53 | 72.0 | 8.92 | 18.61 | 8.35 | 87.8 | 9.57 | 18.77 | 8.93 |
| 3 | - | ✓ | **51.0** | 16.17 | **34.59** | 15.26 | 73.3 | 10.26 | 25.48 | 9.47 | **80.9** | **15.31** | **31.39** | **14.28** |
| 4 | ✓ | ✓ | 54.5 | **17.37** | 33.47 | **16.34** | **64.6** | **12.49** | **27.28** | **11.66** | 84.8 | 14.39 | 29.22 | 13.45 |

## VI. ACKNOWLEDGMENT

This work was kindly supported by Joint Funds of the National Natural Science Foundation of China through grant No. U23A20346, the Key Research and Development Program of Heilongjiang Province through grant No. GNCMSSJH20260012, and the National Key R&D Program of China through grant 2025YFB4712702.